\documentclass[journal,comsoc]{IEEEtran}
\usepackage{amssymb}
\usepackage{amsthm,amssymb}
\usepackage{eucal}
 \usepackage{multicol}
\usepackage{cite}
\usepackage{amsmath,amssymb,amsfonts}
\usepackage{amsmath}

\usepackage{graphicx}
\usepackage[table]{xcolor}
\usepackage{mathtools}  
\usepackage{diffcoeff}  
\usepackage{graphicx}
\usepackage{textcomp}
\usepackage{xcolor}
\usepackage{lettrine}
 \usepackage{multirow}
 \usepackage{amsmath}
 \usepackage{graphicx}
 \usepackage{textgreek}
\usepackage{amssymb}
\usepackage{caption}
\usepackage{subcaption}
\usepackage{authblk}
\usepackage{tikz}
\usetikzlibrary{shapes,arrows}

\usepackage{amsmath}
\newcommand{\RN}[1]{%
  \textup{\uppercase\expandafter{\romannumeral#1}}%
}
\usepackage[ruled,longend]{algorithm2e}
\def\BibTeX{{\rm B\kern-.05em{\sc i\kern-.025em b}\kern-.08em
    T\kern-.1667em\lower.7ex\hbox{E}\kern-.125emX}}
\def\BibTeX{{\rm B\kern-.05em{\sc i\kern-.025em b}\kern-.08em
    T\kern-.1667em\lower.7ex\hbox{E}\kern-.125emX}}
\begin{document}

\title{Machine learning empowered Modulation  detection for OFDM-based signals }
\IEEEoverridecommandlockouts

  \author{\IEEEauthorblockN{\normalsize Ali Pourranjbar,   Georges Kaddoum, Verdier  Assoume Mba,
Sahil Garg,  Satinder Singh.  
}
\vspace{-.75cm}
\thanks{
% Ali Pourranjbar is with the Resilient Machine Learning Institute (ReMI),
% École de Technologie Supérieure (ÉTS), University of Quebec, Montreal,
% QC G1K 9H7, Canada (e-mail: ali.pourranjbar.1@ens.etsmtl.ca).\\
% Georges Kaddoum are with the LaCIME Lab,
% Department of Electrical Engineering, Ecole de Technologie Superieure,
% Montreal, QC H3C 0J9, Canada (e-mail: ibrahim.elleuch.1@ens.etsmtl.ca;
% georges.kaddoum@etsmtl.ca).\\Verdier  Assoume Mba,
% Sahil Garg, and Satinder Singh are with Ultra Intelligence $\&$ Communications Montreal, Canada.  
}
 }
  
\maketitle

\begin{abstract} 
We propose a blind ML-based modulation detection for OFDM-based technologies. Unlike previous works that assume  an ideal environment with precise knowledge of subcarrier count and cyclic prefix location, we consider   blind modulation detection while accounting for realistic environmental parameters and imperfections. \textcolor{black}{
Our approach employs a ResNet network to simultaneously detect the modulation type and accurately locate the cyclic prefix. Specifically, after eliminating the environmental impact from the signal and accurately extracting the OFDM symbols, we convert these symbols into scatter plots.} Due to their unique shapes, these scatter plots are then classified using ResNet. As a result, our proposed modulation classification method can be applied to any OFDM-based technology without prior knowledge of the transmitted signal. We evaluate its performance across various modulation schemes and subcarrier numbers. Simulation results show that our method achieves a modulation detection accuracy exceeding $80\%$ at an SNR of $10$ dB and $95\%$ at an SNR of $25$ dB.

\end{abstract}

 \vspace{-0.5cm}
 \section{Introduction}
Blind modulation classification is primarily explored within the realm of tactical wireless communication, where a wireless node aims to eavesdrop on or jam specific communication links. However, with the emergence of new communication technologies designed to meet high-demand  needs and leverage machine learning (ML), wireless communication can now benefit from a fresh perspective on blind modulation.

Technically, in existing OFDM-based communication technologies, a substantial portion of the frame and subcarrier is allocated to synchronization and pilot signals. This allocation can be reduced by employing blind modulation techniques, thereby enhancing throughput. Blind modulation detection techniques can be broadly categorized into two main groups: likelihood-based (LB) approaches and feature-based methods. Likelihood-based techniques are mainly    based on
maximizing the likelihood function of the received signal
with respect to the unknown modulation classes. Though
LB classifiers are optimal in the Bayesian sense, their
high computational complexity and analytical difficulties in
modeling the unknown parameters often result in   sub-optimal solutions
[6] that affect the overall system performance. 
LB Algorithms utilize precise or approximated probability functions \cite{chung1995likelihood} for addressing the classification issue, which is essentially considered as a multi-hypothesis examination \cite{dobre2005blind}. In the context of non-blind Adaptive Modulation and Coding (AMC), achieving the Maximum Likelihood (ML) solution would be ideal in terms of Correct Classification Rate (CCR). However, it presents two main limitations. Firstly, both exact and approximated ML solutions are typically impractical due to the absence of a straightforward closed-form or the associated computational complexity. Secondly, LB algorithms commonly lack robustness against model discrepancies. In scenarios involving a blind environment and multipath fading channels, LB solutions are evidently disregarded. 

In \cite{a1}, the authors introduced a modulation detection method that utilizes the second-order, fourth-order, and sixth-order cumulants of  consecutive subcarriers to identify modulation types within the QAM group and QPSK.  In \cite{gupta2018blind}, the authors employed the cyclo stationarity and higher-order statistical properties of the received baseband signal for classification. The techniques suggested in \cite{a1} and \cite{gupta2018blind} were applicable in OFDM-based systems, albeit with certain practical limitations. These limitations stem from the unrealistic assumption that there is access to information, such as the location and size of the CP  as well as the minimal carrier offset \cite{a1}, and the restriction to only two types of modulation, QPSK and MSK. Consequently, these methods may not be suitable for OFDM-based technologies employing QAM modulation schemes. \textcolor{black}{In addition, they assume that the modulation of the OFDM symbols is the same for conscutive number of  symbols, which is not a practical assumption for real-world environments.}

FB classifiers use the features extracted from the received
signal to identify its modulation format. ML-based classifiers are counted as  FB classifiers. ML techniques for non-OFDM technologies are well-studied\cite{norolahi2022blind,qiao2022blind,deng2023co,snoap2024deep}. However, OFDM-based communication technologies have unique structures, such as the existence of a CP and sensitivity to carrier frequency offset (CFO), which can cause a loss of orthogonality.
\textcolor{black}{ Some existing studies, such as \cite{kumar2023automatic} and \cite{ren2024ofdm}, propose blind modulation detection but have significant drawbacks like LB classifiers, including their reliance on the assumption that consecutive OFDM symbols use identical modulation schemes}. The study in \cite{kumar2023automatic} proposed using a CNN network with residual blocks for modulation classification of OFDM-based signals using the time domain of the received signal.  The authors in \cite{ren2024ofdm} introduced Cross-SKNet for OFDM modulation in the presence of CFO and IQ imbalance.

% . Although they considered some realistic environmental factors like CFO, they did not account for the possibility of the modulation scheme changing during the stream, assuming instead a fixed scheme for each stream. Additionally, their proposed structure is not suitable for decoding data after modulation detection, as their method does not specify the exact location of the data.
%Some of the machine leanring based such as \cite{yang2024sensing} uses statistical dispersion of
%amplitude (NSDA) and high-order statistics as a feature for modulation detection of the OFDM based signal.

In general, numerous blind modulation techniques have been proposed for non-OFDM-based technologies, making them unsuitable for OFDM-based systems. Existing studies on blind modulation techniques for OFDM typically suffer from assumptions that are impractical in real-world scenarios. These assumptions often include the availability of full Channel State Information (CSI), the known number of subcarriers, and the exact location and size of the CP. Such assumptions are not feasible in practical applications.

% Moreover, some existing works overlook the fact that the modulation scheme of each OFDM symbol can vary. Even among those that consider the possibility of different modulation schemes for each OFDM symbol, they often assume perfect CP location determination and perfect extraction of the OFDM symbols. These idealized conditions do not align with the challenges faced in practical implementations, where such perfect information is rarely available.

\textcolor{black}{In this work, we propose a blind modulation classification method that operates without any prior information and can both extract the OFDM symbol and determine its modulation.  Unlike existing works that assume all OFDM symbols in a stream have uniform modulation, our approach allows the modulation to vary from symbol to symbol, as is the case in practice. }

The rest of the paper is organized as follows: Section II presents the
system model and problem formulation. Section III elaborates on the proposed
blind modulation classification. Simulation
results are exhibited in Section IV, and conclusions are drawn in
Section V.

\vspace{-0.3cm}
%%%%%%%%%%%
\section{OFDM signal model }

We consider an OFDM-based system with $N$ sub-carriers where
the samples of the $m$th OFDM symbol are generated 
using N-point inverse DFT (IDFT), expressed as
\begin{equation}
S_m[k] =\sum_{n=0}^{N-1}  S_m[n]e^{j2\pi n\frac{k}{N}}
\end{equation}
where $S_m[n]$ is the modulated signal of the $m$th OFDM symbol. After taking the IFFT of the signal, the CP with  length  $N_{cp}$ is added to the signal to avoid inter symbol interference and transmitted to the receiver. As a result, the transmitted signal is

\[ s_m[k] = \begin{cases} 
s_m[k + N], & \text{if } -N_{cp} \leq k \leq -1 \\
s_m[k], & \text{if } 0 \leq k \leq N - 1 .
\end{cases}
\]

\noindent Assuming that the sampling rate is $f_s=Nf_{ss}$, where $f_{ss}$ is the sub-carrier spacing, the received OFDM signal after passing through a multi-path fading channel  can be expressed as 
\begin{equation}
\begin{aligned}
 r_m[k] = &e^{j(2\pi (f_{frac}+ f_{int})\frac{k}{Nf_{ss}}+\phi)} \sum_{l=0}^{L-1}  h[l]S_m[k-l-\tau] + w[k],\\ &0\leq k\leq R-1
 \end{aligned}
 \end{equation}

\noindent where $f_{frac}$ and $ f_{int}$ are the fraction and integer parts of the frequency offset \footnote{\textcolor{black}{The fraction part of the carrier offset denotes the proportion of the subcarrier spacing that represents the non-integer component of the frequency difference between the received signal and the local oscillator. In contrast, the integer part of the carrier offset indicates how many times the subcarrier spacing fits into this frequency difference as a whole number.}}, $\phi$ is the phase offset, $\tau$ is the timing offset due to the wireless channel, $w[k]$ is the AWGN with zero mean and variance $\sigma^2_w$, and  $R = N + N_{cp}$ is the received signal length. The channel impulse response (CIR) of the frequency-selective fading channel is denoted by $h[l]$, where $l = 0, 1, \ldots, L - 1$, with $L$ being the length of the CIR, providing $N_{cp} \geq L$.

Traditionally, in the current setup, the transmitter and
receiver exchange pilot and synchronization signals to estimate channel characteristics, CFO, and noise power.  
Thus, upon receiving the signal, the receiver compensates for the effects of physical distortion and proceeds to detect the modulation or modulations utilized in the current symbols. 
It's notable that the modulation and coding rate   used are stored in the control channel data, which is exchanged after channel estimation. 
Given that the receiver and transmitter adhere to specific structures and standards, the receiver can extract the information necessary to determine the modulation and subcarrier indexing of each incoming data stream. In blind demodulation, the receiver lacks any prior knowledge about the transmitter, and the exchange of pilot signals and synchronization signals is not feasible. Therefore, it becomes necessary to adopt a method capable of detecting the modulation without relying on any prior information about the transmitter.
 
\vspace{-0.3cm}
 \begin{figure}[t]
    \centering
    \includegraphics[width=0.5\textwidth, height=0.25\textwidth]{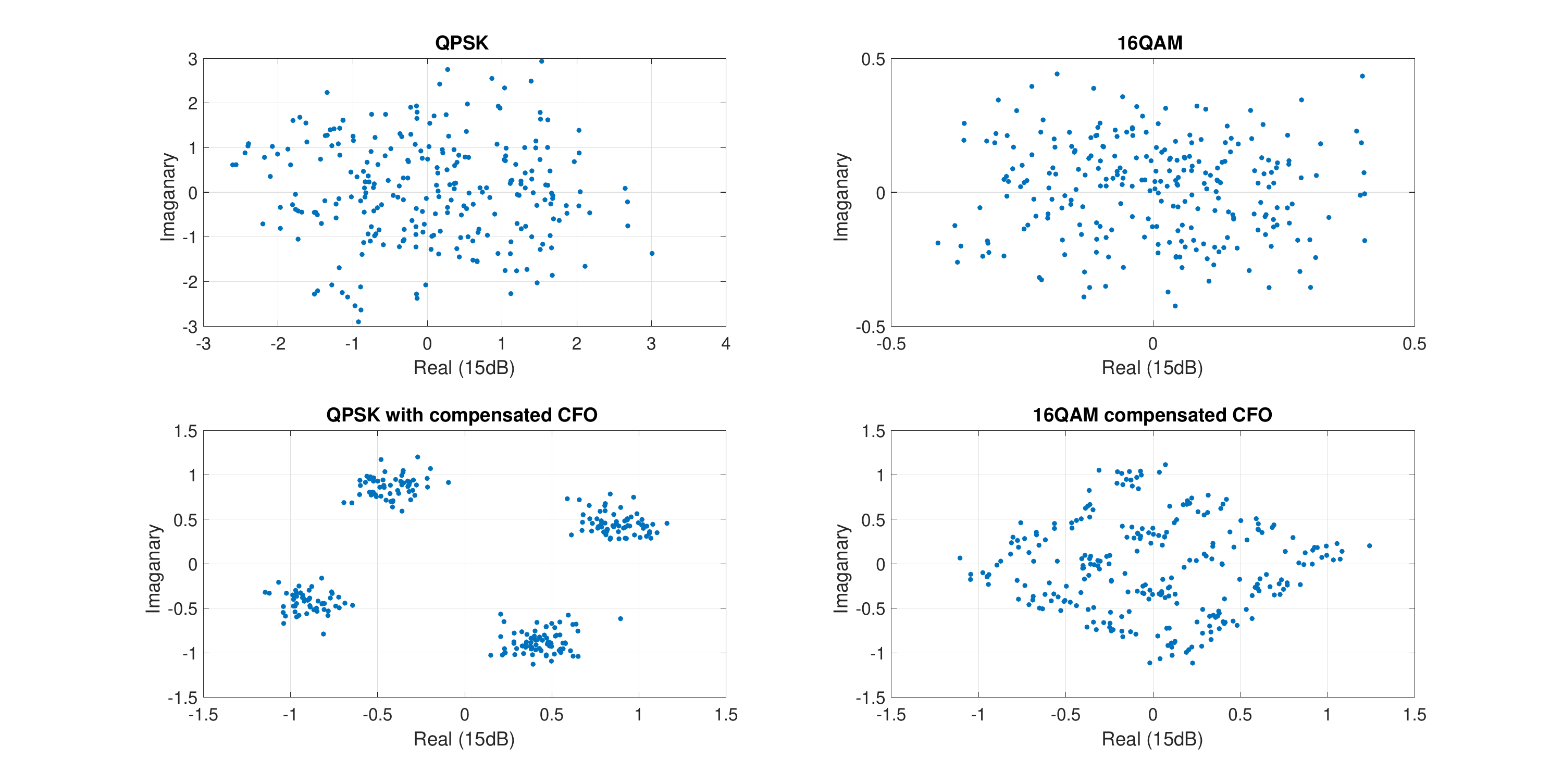}
    \caption{Constellation structure before and after the effects of CFO under multi-path fading and noise.}
    \label{fig1}
\end{figure}
 
\section{Blind modulation classification }
In traditional OFDM symbol detection, when the transmitter and receiver exchange information, the receiver follows several steps to demodulate an OFDM symbol. Firstly, it obtains necessary information, such as the number of subcarriers and CFO, through the primary and secondary synchronization signals. Then, by decoding the control channel, it determines the location and size of each  CP, as well as the utilized modulation. In the blind modulation technique, we follow the same process until the utilized modulation is obtained. As mentioned above, the first step is to find the number of sub-carriers.
\vspace{-0.3cm}
\subsection{Number of sub-carriers and subcarrier spacing}
To ascertain the number of subcarriers, one can leverage the cyclical nature inherent in OFDM symbols. To this end, the received signal is correlated with a delayed version of itself, shifted by various intervals. The position where the signal exhibits its peak yields the sub-carrier count. Subsequently, the sub-carrier spacing $\Delta f$ is derived by dividing the monitored bandwidth under consideration by the determined number of sub-carriers. Once the subcarrier count 
  is established, the next step involves pinpointing the  CP  location and calculating the distance from the received signal to the first CP location. Here again, the cyclic characteristic of OFDM signals proves instrumental.
\begin{equation}
(r \star y)[p] = \sum_{n=0}^{Z-1} r[n] \cdot y[n-p]
\end{equation}
where:
\begin{equation}
y[n] = \begin{cases} r[n-d] & \text{if } d \leq n < Z+d \\ 0 & \text{otherwise} \end{cases}
\end{equation}
$r[n]$ is the original signal,  
  $Z$ is the length of the received signal,    
  $d$   is the delay, and    
  $p$   ranges from   0   to   $P$,   where $P$ is the maximum possible number of OFDM sub-carrier which is set to $2048$ as in the case of LTE.
 
\vspace{-0.3cm}
% \begin{figure}[t]
%     \centering
%     \includegraphics[width=0.5\textwidth, height=0.3\textwidth]{fig2.eps}
%     \caption{Constellation structure after removing fraction part of   CFO for 10dB and 15dB.}
%     \label{fig2}
% \end{figure}
\subsection{Finding the CP size and coarse of timing. }

In determining the CP size and location, we exploit the fact that the CP is a repetition of the last part of each OFDM symbol. By correlating the CP  to the last segment of the OFDM symbol, the correlation yields its peak value, aiding in CP size and location identification, along with the distance to the first CP.

In our approach, we consider multiple CP sizes, typically ranging from $6\%$ to $15\%$ of the total number of subcarriers. Assuming the start of the symbol aligns with the first sample of the received signal, we correlate the CP to the main part over five or six consecutive OFDM symbols for each considered CP size.We repeat this process for all CP sizes, retaining the indices of all CP sizes with the highest correlation values. Subsequently, we shift one sample and iterate the correlation process, continuing until half the number of Fast Fourier Transform (FFT) samples is reached. We record the maximum correlation value for each shifted sample. Ultimately, the sample exhibiting the highest correlation value denotes the proper shift(i.e. coarse timing), and its corresponding index indicates the desired CP size. It's important to note that due to the influence of noise and channel fading, the  estimated CP value may slightly deviate from the actual one. However, this deviation is rectified in the subsequent modulation phase, which we will elaborate on shortly.
\vspace{-0.3cm}
\subsection{CFO }

After determining the approximate CP size and shift value, we proceed to estimate the fraction part of the CFO \textcolor{black}{($f_{frac}$). Firstly, we calculate the phase difference between a detected CP and its replica using cross-correlation. To convert this phase difference into frequency, we divide the calculated phase by $2\pi  \Delta f N$}. The resulting value provides an estimate of the CFO. Once the fraction part of the CFO is estimated, we subtract it from the main received signal, effectively removing its effect. In Fig. \ref{fig1}, QPSK and 16 QAM modulations are presented before and after removing the fraction part of the CFO under multipath fading with an SNR of $15$dB. Fig. \ref{fig1} shows that after removing the CFO, scatter plots of the OFDM symbols have their own unique shapes that can be used to distinguish one modulation from another. Thus, we utilize these scatter plots for  modulation classification, which is detailed in the next section.
\vspace{-0.3cm}
\subsection{Modulation }

Once the approximate CP size and the distance to the first CP  are determined and the CFO is removed, we can proceed with modulation classification. Initially, we remove the CP to isolate the OFDM symbol. Then, we perform Fast Fourier Transform (FFT) on the OFDM symbol using the determined number of subcarriers. The resulting symbol can be utilized for modulation detection. Since channel detection and equalization are not performed, mathematical-based methods, like Maximum Likelihood (ML) cannot be used for determining the modulation type. Therefore, we propose employing ML methods for modulation type detection in this stage. The proposed modulation classification consists of two parts: training and testing. During the training phase, the network is trained for modulation detection. In the testing phase, the trained network is evaluated using the test set.

 \begin{figure}[t]
    \centering
    \includegraphics[width=0.5\textwidth, height=0.2\textwidth]{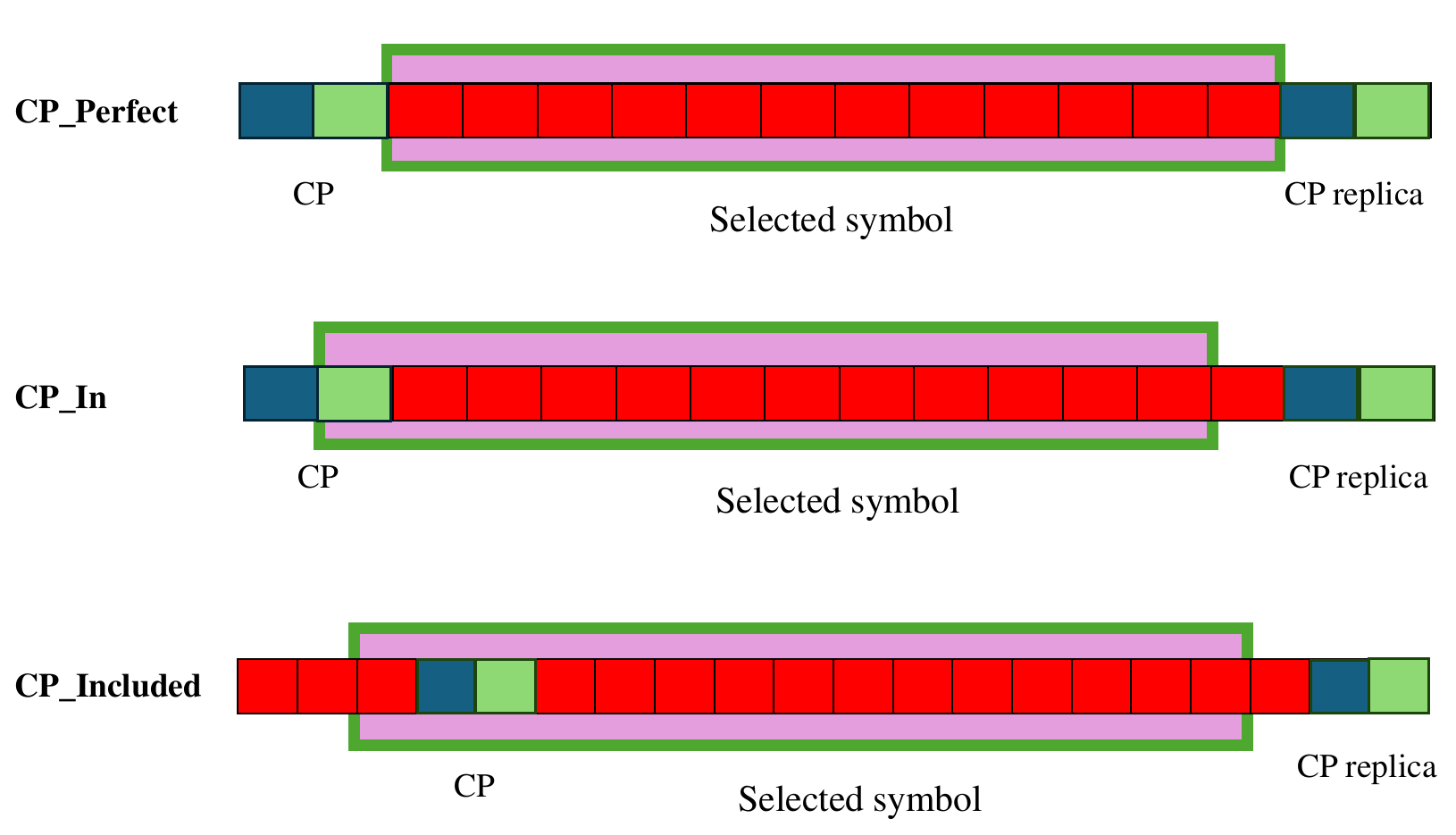}
    \caption{CP scenarios.}
    \label{CP_}
\end{figure}

 \vspace{-0.3cm}

%%%%%%%%%%%%%%%
 \subsection{Training}
 \subsubsection{Data preparation}

Before delving into the training process, it is essential to understand the data preparation process. To generate the dataset, we focus on the standard modulations commonly used in OFDM-based technologies: BPSK, QPSK, and QAM group. The QAM group encompasses 16 QAM, 64 QAM, 256 QAM, and 1024 QAM. Thus, the trained network should be provided with data from these six modulation type. Furthermore, considering that the estimation of the CP location and size can be affected by noise, we need to incorporate scenarios where the CP location and size are determined with some error in the training data generation process. Therefore, we define three distinct scenarios for our model, as depicted in Fig. \ref{CP_}, elaborated on next.
%%%%%%%%%%%

Specifically, the elaborated scenarios for the data preparation process are detailed as follows:
\begin{itemize}
\item {The first scenario involves accurately determining the location of the CP, resulting in the proper separation of the OFDM symbol from the received signal. This scenario is named "CP$\_$perfect".}

\item {In the second scenario, the CP location is correctly identified, but the CP is included within the symbol itself, with the first sample of the symbol coinciding with the CP. This scenario is denoted as "CP$\_$in."}

\item {The last scenario occurs when the CP is included in separate symbols, but the symbol does not begin with a CP. This scenario is labeled as "CP$\_$included."}
\end{itemize}
These scenarios encompass the various conditions that may arise due to noise and other factors during the CP location and separation process. Moreover, the generated dataset covers the following range of factors to ensure comprehensive training:
\begin{itemize}
\item {
Number of OFDM subcarriers: Varies between 128 and 2048, covering different system configurations.}
\item {
Carrier Frequency Offset (CFO): Ranging from 100 to 500 ppm, to account for varying levels of signal distortion.}
\item {
Noise Power: Ranging from 10 to 25 dB, reflecting different levels of background noise.}
\item {
Fading Channel Model: Utilized the SUI I model to simulate real-world channel conditions.}
\end{itemize}

Additionally, to simulate the compensation process for  CFO, a random frequency is initially generated. Subsequently, the fractional part of this frequency is compensated, mirroring the real-world compensation procedure outlined earlier. This  dataset allows the network to learn and generalize across various system configurations and environmental conditions.

 \begin{figure}[t]
    \centering
    \includegraphics[width=0.45\textwidth, height=0.2 \textwidth]{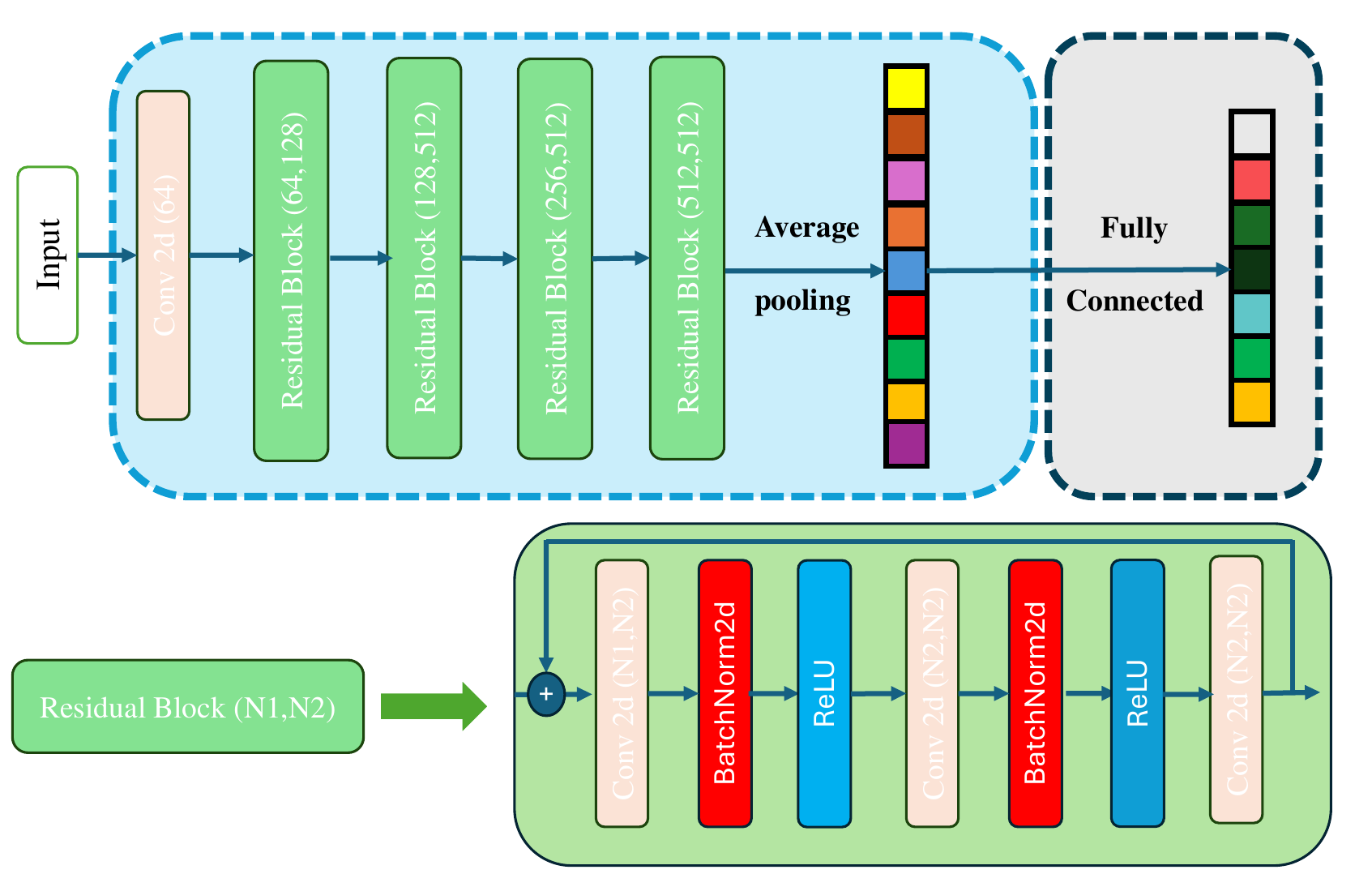}
    \caption{Network model architecture.}
    \label{fig4}
\end{figure}
Following completing the data generation process, the next step is to utiliz  the data for training. Since the generated data is in the time domain, \textcolor{black}{we need to perform a Fast Fourier Transform (FFT) on the sample data and convert it to the frequency domain (The FFT size is equal to N)}. Once the FFT is implemented, the resulting signal is converted into 2D scatter plots of 400x400 pixels. In these scatter plots, the imaginary part of the FFT is represented on the Y-axis and the real part on the X-axis. This conversion to scatter plots is motivated by the unique shape characteristics of the different modulations, which can be effectively used for differentiation.

Each scatter plot serves as an input, and its corresponding modulation type is considered the target label for training. The labels are specified as follows:
$[$\textnormal{CP$\_$in} = 0, \textnormal{CP$\_$included} =1, QPSK = 2, BPSK = 3, 16QAM = 4, 64QAM = 5, 256QAM = 6, 1024QAM = 7$]$.

To perform the classification, a Convolutional Neural Network (CNN)  is employed due to its effectiveness in handling image data. Specifically, Residual Networks (ResNets) are utilized, leveraging shortcut connections to mitigate the vanishing gradient problem and enabling the training of very deep architectures with numerous  layers. The overall network topology is depicted in Fig. \ref{fig4}. During the training process, the generated scatter plots of FFT data are fed into the network, and the corresponding labels are provided as targets. The cross-entropy function is chosen as the loss function.

 \begin{figure}[t]
    \centering
    \includegraphics[width=0.5\textwidth, height=0.2 \textwidth]{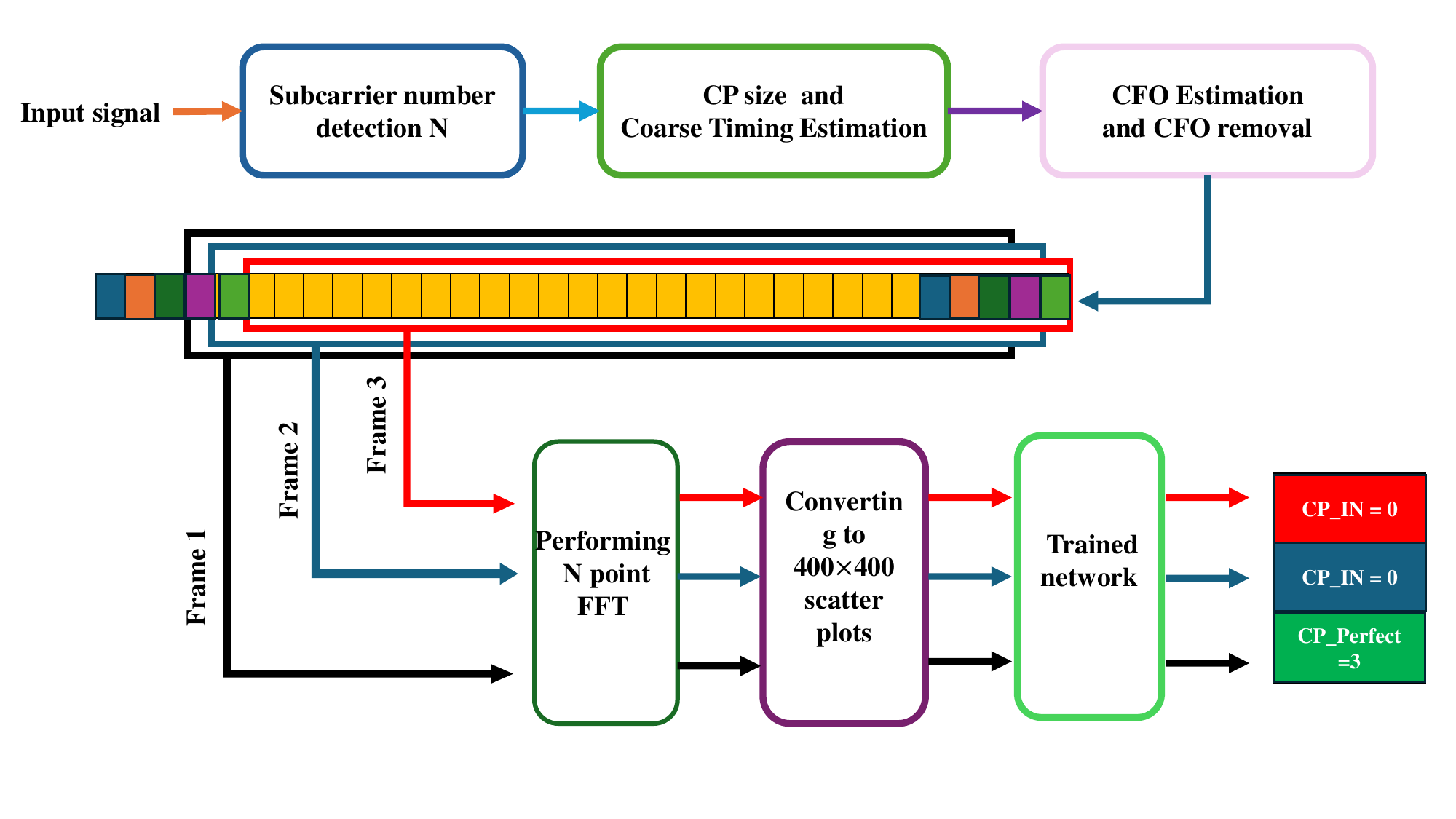}
    \caption{Test process.}
    \label{Tproces}
\end{figure}
\vspace{-0.3cm}
\subsection{Testing}

During the testing process, a stream of received OFDM data undergoes the aforementioned steps, as depicted in Fig. \ref{Tproces}. Initially, the number of subcarriers in the received samples is determined, followed by the localization of the CP and \textcolor{black}{number of samples to the first OFDM symbol}. The CFO is then estimated and removed from the data. Due to potential errors in CP detection, the number of OFDM frames is provided to the network for each OFDM symbol. \textcolor{black}{In essence, the initial OFDM frame begins halfway through the first detected CP ($N_{CP}/2$), and the final considered OFDM frame starts at $3N_{CP}/2$.}

The OFDM frame begins halfway through the estimated CP, extending to the number of samples equivalent to half of the CP after the estimated CP location. Given that during training, "CP$\_$In" and "CP$\_$Included" are treated as separate labels, the trained network distinguishes between these scenarios and enhances CP detection accuracy. The FFT of the considered frames are computed and converted into scatter plots, which are then inputted to the trained network. Subsequently, the network generates labels indicating the status of each  frame separately. Once the modulation is detected, the index of the modulation is added to the calculated shift, facilitating the detection of the next OFDM symbol in the received data stream. This iterative process enables accurate modulation classification despite potential CP detection errors.

%%%%%%%%%%%%%%%%%%%%

% \begin{figure}
%     \centering
%     \includegraphics[width=0.5\textwidth, height=0.25\textwidth]{Diagram.pdf}
%     \caption{Caption}
%     \label{fig4}
% \end{figure}
\vspace{-0.2cm}
\section{Simulation Results}
In this section, we evaluate our proposed method by testing it on  LTE data generated by the Matlab waveform generator toolbox. To this end, we generate a wide range of OFDM signals with different sub carrier numbers and signal to noise ratios (SNR), where each signal is affected by a random CFO from 100 to 500 PPM. \textcolor{black}{Moreover, the modulation of each OFDM symbol is selected randomly and changes with every symbol.}  The central carrier is set to $2$GHz.
\begin{table}[t]
\centering
\tiny
\begin{tabular}{ccccccccc}
\multicolumn{1}{c}{10dB} & \multicolumn{1}{c}{\cellcolor{gray!30}0} & \multicolumn{1}{c}{\cellcolor{gray!30}1} & \multicolumn{1}{c}{\cellcolor{gray!30}QPSK} & \multicolumn{1}{c}{\cellcolor{gray!30}BPSK} & \multicolumn{1}{c}{\cellcolor{gray!30}16 QAM} & \multicolumn{1}{c}{\cellcolor{gray!30}64 QAM} & \multicolumn{1}{c}{\cellcolor{gray!30}256 QAM} & \multicolumn{1}{c}{\cellcolor{gray!30}1024 QAM} \\
\cellcolor{gray!30}0 & \cellcolor{green!45} 59.01 & \cellcolor{red!50} 38.07 & 0 & 0 & \cellcolor{red!5} 0.55 & \cellcolor{red!5} 1.48 & 0.76 & 0.13 \\
\cellcolor{gray!30}1 & \cellcolor{red!15} 5.21 & \cellcolor{green!67} 93.39 & 0 & 0 & \cellcolor{red!4} 0.05 & \cellcolor{red!4} 0.35 & 0 & 0 \\
\cellcolor{gray!30}QPSK & 0 & 0 & \cellcolor{green!100} 99.95 & 0 & \cellcolor{red!4} 0.05 & 0 & 0 & 0 \\
\cellcolor{gray!30}BPSK & 0 & 0 & 0 & \cellcolor{green!100} 100 & 0 & 0 & 0 & 0 \\
\cellcolor{gray!30}16 QAM & 0.19 & 0.16 & 0 & 0 & \cellcolor{green!45} 96.15 & \cellcolor{red!5} 3.43 & 0.06 & 0 \\
\cellcolor{gray!30}64 QAM & 0.50 & 0.31 & 0 & 0 & 1.81 & \cellcolor{green!97} 66.90 & \cellcolor{red!4} 25.94 & 4.53 \\
\cellcolor{gray!30}256 QAM & 0.06 & 0.23 & 0 & 0 & 0 & \cellcolor{red!15} 23.73 & \cellcolor{green!67} 59.50 & \cellcolor{red!25} 16.47 \\
\cellcolor{gray!57}1024 QAM & 0 & 0 & 0 & 0 & 0 & 8.60 & \cellcolor{red!25} 31.86 & \cellcolor{green!37} 59.55 \\
\multicolumn{1}{c}{15dB} & \multicolumn{1}{c}{\cellcolor{gray!30}0} & \multicolumn{1}{c}{\cellcolor{gray!30}1} & \multicolumn{1}{c}{\cellcolor{gray!30}QPSK} & \multicolumn{1}{c}{\cellcolor{gray!30}BPSK} & \multicolumn{1}{c}{\cellcolor{gray!30}16 QAM} & \multicolumn{1}{c}{\cellcolor{gray!30}64 QAM} & \multicolumn{1}{c}{\cellcolor{gray!30}256 QAM} & \multicolumn{1}{c}{\cellcolor{gray!30}1024 QAM} \\
\cellcolor{gray!30}0 & \cellcolor{green!45} 64.01 & \cellcolor{green!37} 34.20 & 0 & 0 & \cellcolor{red!5} 0.12 & \cellcolor{red!5} 1.35 & \cellcolor{red!5} 0.30 & 0.02 \\
\cellcolor{gray!30}1 & \cellcolor{red!15} 9.49 & \cellcolor{green!67} 89.19 & 0 & 0 & \cellcolor{red!4} 0.18 & \cellcolor{red!4} 0.72 & \cellcolor{red!4} 0.41 & 0 \\
\cellcolor{gray!30}QPSK & 0 & 0 & \cellcolor{green!100} 100 & 0 & 0 & 0 & 0 & 0 \\
\cellcolor{gray!30}BPSK & 0 & 0 & 0 & \cellcolor{green!100} 100 & 0 & 0 & 0 & 0 \\
\cellcolor{gray!30}16 QAM & 0 & 0 & 0 & 0 & \cellcolor{green!100} 100 & 0 & 0 & 0 \\
\cellcolor{gray!30}64 QAM & \cellcolor{red!5} 0.17 & 0 & 0 & 0 & 0 & \cellcolor{green!97} 96.94 & \cellcolor{red!25} 2.86 & 0.03 \\
\cellcolor{gray!30}256 QAM & \cellcolor{red!25} 0.09 & 0 & 0 & 0 & 0 & \cellcolor{red!25} 5.93 & \cellcolor{green!67} 80.84 & \cellcolor{red!50} 13.14 \\
\cellcolor{gray!57}1024 QAM & 0 & 0 & 0 & 0 & 0 & \cellcolor{red!50} 0.28 & \cellcolor{red!37} 38.51 & \cellcolor{green!45} 61.20 \\
\multicolumn{1}{c}{20dB} & \multicolumn{1}{c}{\cellcolor{gray!30}0} & \multicolumn{1}{c}{\cellcolor{gray!30}1} & \multicolumn{1}{c}{\cellcolor{gray!30}QPSK} & \multicolumn{1}{c}{\cellcolor{gray!30}BPSK} & \multicolumn{1}{c}{\cellcolor{gray!30}16 QAM} & \multicolumn{1}{c}{\cellcolor{gray!30}64 QAM} & \multicolumn{1}{c}{\cellcolor{gray!30}256 QAM} & \multicolumn{1}{c}{\cellcolor{gray!30}1024 QAM} \\
\cellcolor{gray!30}0 & \cellcolor{green!45} 99.61 & 0 & 0 & 0 & \cellcolor{red!5} 0.26 & 0.14 & 0 \\
\cellcolor{gray!30}1 & \cellcolor{red!15} 9.99 & \cellcolor{green!67} 88.30 & 0 & 0 & \cellcolor{red!4} 0.21 & 1.10 & 0.40 & 0 \\
\cellcolor{gray!30}QPSK & 0 & 0 & \cellcolor{green!67} 99.97 & \cellcolor{red!4} 0.03 & 0 & 0 & 0 & 0 \\
\cellcolor{gray!30}BPSK & 0 & 0 & 0 & \cellcolor{green!100} 100 & 0 & 0 & 0 & 0 \\
\cellcolor{gray!30}16 QAM & 0 & 0 & 0 & 0 & \cellcolor{green!100} 100 & 0 & 0 & 0 \\
\cellcolor{gray!30}64 QAM & \cellcolor{red!5} 0.07 & 0 & 0 & 0 & 0 & \cellcolor{green!97} 99.93 & 0 & 0 \\
\cellcolor{gray!30}256 QAM & 0 & 0 & 0 & 0 & 0 & 0 & \cellcolor{green!45} 99.69 & \cellcolor{red!5} 0.31 \\
\cellcolor{gray!57}1024 QAM & 0 & 0 & 0 & 0 & 0 & 0 &\cellcolor{red!50} 31.49  &  \cellcolor{green!45} 68.51 \\
\end{tabular}
\caption{Confusion tables for modulation classification.}
\label{tab-20dB}
\end{table}
% %%%%%%%%%%%%%%%%%%%%%%% 

% \begin{table}[t]
% \centering
% \tiny
% \begin{tabular}{cccccccc}
% \multicolumn{1}{c}{Label} & \multicolumn{1}{c}{\cellcolor{gray!30}0} & \multicolumn{1}{c}{\cellcolor{gray!30}1} & \multicolumn{1}{c}{\cellcolor{gray!30}2} & \multicolumn{1}{c}{\cellcolor{gray!30}3} & \multicolumn{1}{c}{\cellcolor{gray!30}4} & \multicolumn{1}{c}{\cellcolor{gray!30}5} & \multicolumn{1}{c}{\cellcolor{gray!30}6} \\
% \cellcolor{gray!30}0 & 0 & 9.98e-01 & 0 & 0 & 1.31e-03 & 8.21e-04 & 0 \\
% \cellcolor{gray!30}1 & 0 & 1 & 0 & 0 & 0 & 0 & 0 \\
% \cellcolor{gray!30}2 & 0 & 5.12e-01 & 6.61e-03 & 6.01e-03 & 3.28e-02 & 5.17e-02 & 1.96e-01 \\
% \cellcolor{gray!30}3 & 0 & 3.02e-01 & 3.00e-04 & 5.52e-01 & 5.56e-03 & 6.46e-03 & 3.98e-02 \\
% \cellcolor{gray!30}4 & 0 & 9.07e-01 & 0 & 5.88e-04 & 4.50e-02 & 2.15e-02 & 2.04e-02 \\
% \cellcolor{gray!30}5 & 0 & 9.76e-01 & 0 & 3.62e-04 & 1.63e-03 & 1.41e-02 & 5.97e-03 \\
% \cellcolor{gray!30}6 & 0 & 9.96e-01 & 0 & 0 & 0 & 0 & 3.59e-03 \\
% \cellcolor{gray!30}7 & 0 & 1 & 0 & 0 & 0 & 0 & 0 \\
% \end{tabular}
% \caption{Your caption here.}
% \label{your-label-here}
% \end{table}
% %%%%%%%%%%%%%%%%%%%%
In Table \ref{tab-20dB}, the confusion matrices for the modulation classification task are presented when the SNR  of 10 dB, 15 dB, and 20 dB. For all the considered scenarios, the precision of QPSK, BPSK, and 16 QAM approaches 1, indicating the network’s high performance. The accuracy for 64 QAM starts at 97 $\%$ and reaches 99.98 $\%$ when the SNR is 20 dB. For 256 QAM, it starts at 59.50 $\%$, reaches 80.84 $\%$ at 15 dB, and ends at 99.69 $\%$. A similar trend holds for  1024 QAM. In high-order QAM, there are many points, which under noisy environments, cause the  picture  to appear  clustered, causeing performance degradation, especially when the number of subcarriers is high. This can be seen for 1024 QAM, where there is a frequent confusion between 256 QAM and 1024 QAM due to the resemblance of their constellations in the presence of noise. Distinguishing between 256 QAM and 1024 QAM in a noisy environment is particularly challenging, even for a human observer. Employing a higher SNR or noise-canceling methods significantly enhances accuracy. It is important to emphasize that this accuracy pertains to sub-modulation, and the detection accuracy between different modulations is nearly 99 $\%$. Labeling CP${IN}$ and CP${Included}$ is crucial for accurately determining the correct positioning of the symbols. Ensuring that the network does not confuse them with actual modulations is essential. If there is confusion between CP${IN}$ and CP${Included}$, it is not significant, as in both cases, they are disregarded, and the process moves on to the next step. Conclusively, the results show that the accuracy of the proposed network significantly increases when the SNR is raised to 15 dB and reaches a very high accuracy when it reaches 20 dB, demonstrating that the trained network is highly accurate.

\begin{figure}[t]
    \centering
    \includegraphics[width=0.5\textwidth, height=0.25\textwidth]{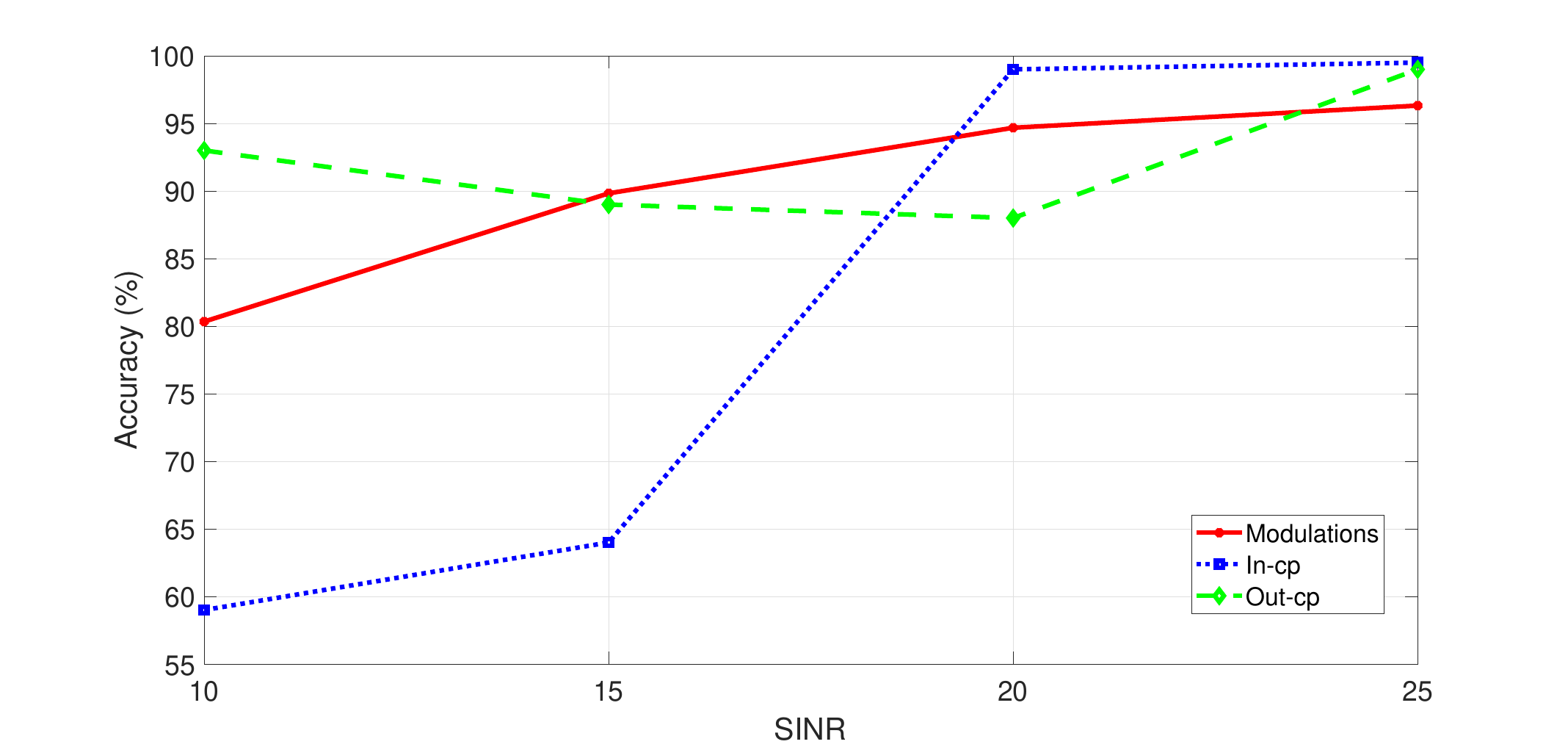}
   
    \caption{Detection accuracy as a function  of the SNR.}
    \label{Accuracy}
\end{figure}

In Fig. \ref{Accuracy}, the average accuracy is presented as the SNR ranges from 10 dB to 25 dB. The findings show that, for the modulations and in-CP scenarios, the accuracy improves with higher SNR. Additionally, in the out-CP scenario, accuracy initially decreases with increasing SINR and then rises again. This occurs because, due to the similarity between in-CP and out-CP, most in-CP and out-CP are identified as out-CP, leading to an increase in the out-CP accuracy. However, as SINR continues to increase, more in-CP symbols are correctly identified, reducing the number of symbols misclassified as out-CP.

% In Fig. \ref{Accuracy}, we present the accuracy of detection as a function of different subcarriers considering SNR=15 dB.
 \begin{figure}[t]
    \centering
    \includegraphics[width=0.5\textwidth, height=0.25\textwidth]{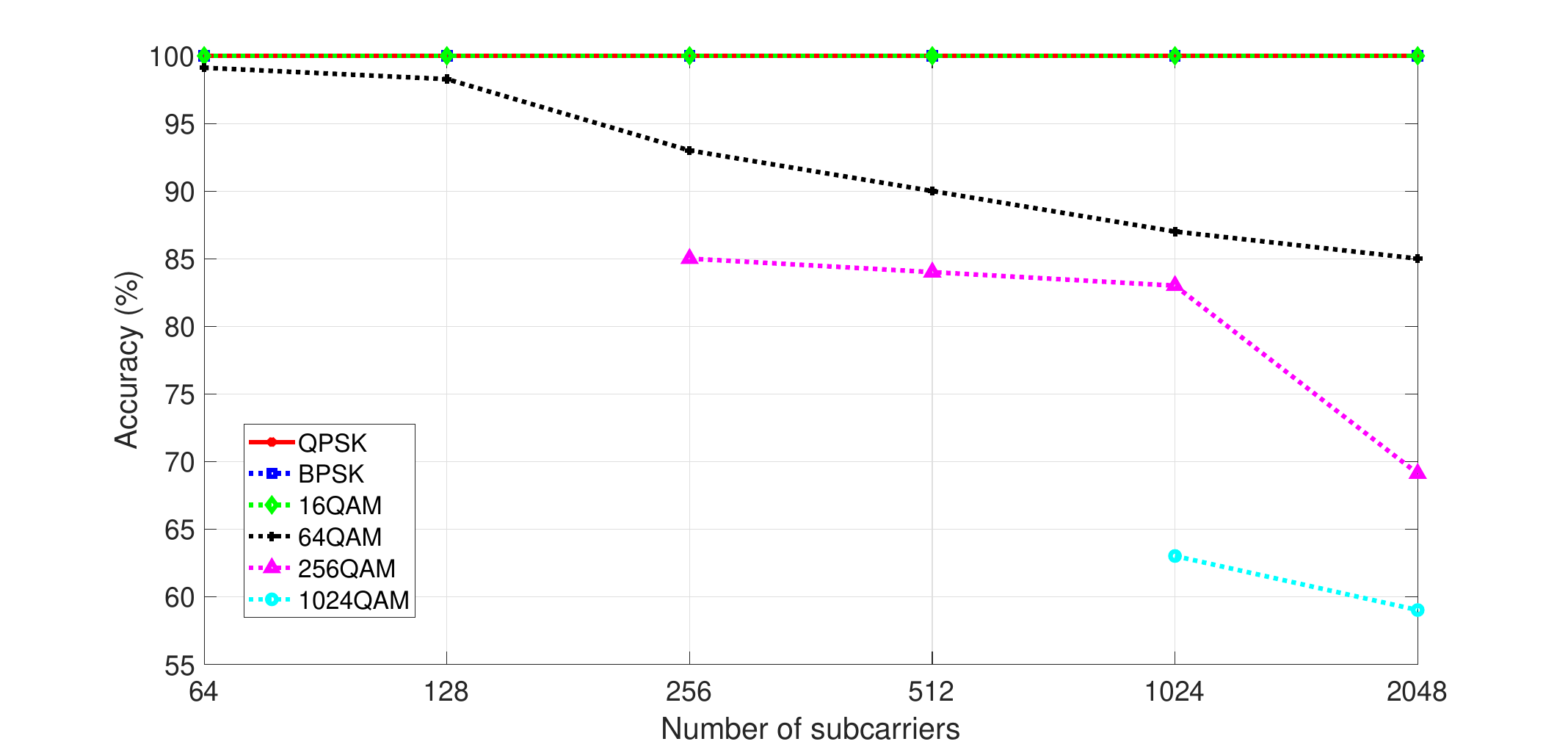}
   
    \caption{   Detection accuracy as a function of the number of subcarriers.}
    \label{Accuracy-sub}
\end{figure}

In Fig. \ref{Accuracy-sub}, We show the detection accuracy as a function of the number of subcarriers. The results indicate that the detection accuracy of the proposed method remains around 100 for BPSK, QPSK, and 16QAM due to the distinctiveness of these modulation shapes, which are preserved even with a high number of subcarriers. However, for 64QAM, 256QAM, and 1024QAM, the detection accuracy decreases as the number of subcarriers increases. This is because an increased number of subcarriers results in more dots in the constellation diagram, making the symbol shapes resemble a random scatter plot rather than a modulated I/Q signal.
% \textcolor{red}{I have to present results for both cp location finding and modulation}
\vspace{-0.2cm}
\section{Conclusion} 
 In this paper, we proposed a blind ML-based modulation detection method for OFDM-based systems. To ensure practicality in real-world environments, we considered realistic conditions and imperfections. Additionally, we thoroughly examined the necessary steps for modulation detection in existing systems, accounting for full knowledge of the environment, and introduced an equivalent substitution for the blind version. \textcolor{black}{Our proposed method was extensively evaluated through simulations, considering modulation changes symbol by symbol, SNR levels, and the number of subcarriers}. The simulation results demonstrated that, across all considered scenarios, the average detection performance of our method exceeds $80\%$ at an SNR of $10$ dB and surpasses $95\%$ when the SNR is $25$ dB.

\vspace{-0.4cm}
\appendices

\bibliographystyle{IEEEtran}
 
\bibliography{references}

\begin{thebibliography}{10}
\providecommand{\url}[1]{#1}
\csname url@samestyle\endcsname
\providecommand{\newblock}{\relax}
\providecommand{\bibinfo}[2]{#2}
\providecommand{\BIBentrySTDinterwordspacing}{\spaceskip=0pt\relax}
\providecommand{\BIBentryALTinterwordstretchfactor}{4}
\providecommand{\BIBentryALTinterwordspacing}{\spaceskip=\fontdimen2\font plus
\BIBentryALTinterwordstretchfactor\fontdimen3\font minus \fontdimen4\font\relax}
\providecommand{\BIBforeignlanguage}[2]{{%
\expandafter\ifx\csname l@#1\endcsname\relax
\typeout{** WARNING: IEEEtran.bst: No hyphenation pattern has been}%
\typeout{** loaded for the language `#1'. Using the pattern for}%
\typeout{** the default language instead.}%
\else
\language=\csname l@#1\endcsname
\fi
#2}}
\providecommand{\BIBdecl}{\relax}
\BIBdecl

\bibitem{chung1995likelihood}
H.~Chung-Yu, ``Likelihood methods for mpsk modulation classification,'' \emph{IEEE Trans. Commun.}, vol.~43, pp. 189--193, 1995.

\bibitem{dobre2005blind}
O.~A. Dobre, A.~Abdi, Y.~Bar-Ness, and W.~Su, ``Blind modulation classification: a concept whose time has come,'' in \emph{IEEE/Sarnoff Symposium on Advances in Wired and Wireless Communication, 2005.}\hskip 1em plus 0.5em minus 0.4em\relax IEEE, 2005, pp. 223--228.

\bibitem{a1}
M.~Liu, F.~Guo, Y.~Chen, and N.~Zhao, ``Blind modulation classification for ofdm in the presence of carrier frequency offsets,'' pp. 4683--4688, 2023.

\bibitem{gupta2018blind}
R.~Gupta, S.~Majhi, and O.~A. Dobre, ``Blind modulation classification of different variants of qpsk and 8-psk for multiple-antenna systems with transmission impairments,'' in \emph{2018 IEEE 88th Vehicular Technology Conference (VTC-Fall)}.\hskip 1em plus 0.5em minus 0.4em\relax IEEE, 2018, pp. 1--5.

\bibitem{norolahi2022blind}
J.~Norolahi, M.~Mehrnia, and P.~Azmi, ``Blind modulation classification via combined machine learning and signal feature extraction,'' in \emph{2021 International Seminar on Machine Learning, Optimization, and Data Science (ISMODE)}.\hskip 1em plus 0.5em minus 0.4em\relax IEEE, 2022, pp. 266--271.

\bibitem{qiao2022blind}
J.~Qiao, W.~Chen, J.~Chen, and B.~Ai, ``Blind modulation classification under uncertain noise conditions: A multitask learning approach,'' \emph{IEEE Communications Letters}, vol.~26, no.~5, pp. 1027--1031, 2022.

\bibitem{deng2023co}
W.~Deng, X.~Wang, and Z.~Huang, ``Co-channel multi-user modulation classification using data-driven blind signal separation,'' \emph{IEEE Internet of Things Journal}, 2023.

\bibitem{snoap2024deep}
J.~A. Snoap, D.~C. Popescu, and C.~M. Spooner, ``Deep-learning-based classifier with custom feature-extraction layers for digitally modulated signals,'' \emph{IEEE Transactions on Broadcasting}, 2024.

\bibitem{kumar2023automatic}
A.~Kumar, K.~K. Srinivas, and S.~Majhi, ``Automatic modulation classification for adaptive ofdm systems using convolutional neural networks with residual learning,'' \emph{IEEE Access}, 2023.

\bibitem{ren2024ofdm}
B.~Ren, K.~C. Teh, H.~An, and E.~Gunawan, ``Ofdm modulation classification using cross-sknet with blind iq imbalance and carrier frequency offset compensation,'' \emph{IEEE Transactions on Vehicular Technology}, 2024.

\end{thebibliography}
%\bibliography{ref}
\end{document}